\documentclass[sn-mathphys,Numbered]{sn-jnl}

\usepackage{geometry}
\usepackage{graphicx}%
\usepackage{comment}
\usepackage{makecell}
\usepackage{multirow}%
\usepackage{amsmath,amssymb,amsfonts}%
\usepackage{amsthm}%
\usepackage{mathrsfs}%
\usepackage[title]{appendix}%
\usepackage{xcolor}%
\usepackage{textcomp}%
\usepackage{manyfoot}%
\usepackage{booktabs}%
\usepackage{algorithm}%
\usepackage{algorithmicx}%
\usepackage{algpseudocode}%
\usepackage{algorithm}%
\usepackage{listings}%
\usepackage{hyperref}

\newcommand{\bx}{\mathbf{x}}
\newcommand{\ba}{\mathbf{a}}
\newcommand{\bs}{\mathbf{s}}
\newcommand{\by}{\mathbf{y}}
\newcommand{\bp}{\mathbf{p}}

\begin{document}

\title{Dynamic feature selection in medical predictive monitoring by reinforcement learning}

\author[1]{\fnm{Yutong} \sur{Chen} \dgr{M.Sc. candidate}}\email{chenyt19@tsinghua.org.cn}

\author*[3]{\pfx{Dr.} \fnm{Jiandong} \sur{Gao} \dgr{Ph.D.}}\email{jdgao@tsinghua.edu.cn}
\equalcont{Corresponding Author}

\author*[4]{\pfx{Dr.} \fnm{Ji} \sur{Wu} \dgr{Ph.D.}}\email{wuji\_ee@tsinghua.edu.cn}
\equalcont{Corresponding Author}

\affil[1,2,3,4]{\orgdiv{Department of Eletronic Engineering}, \orgname{Tsinghua University}, \orgaddress{\city{Beijing}, \country{China}}}

\affil[3,4]{\orgdiv{Center for Big Data and Clinical Research, Institute for Precision Medicine}, \orgname{Tsinghua University}, \orgaddress{\city{Beijing}, \country{China}}}

\keywords{time-series, feature cost, reinforcement learning, feature selection}

\abstract{
   In this paper, we investigate dynamic feature selection within multivariate time-series scenario, a common occurrence in clinical prediction monitoring where each feature corresponds to a bio-test result. Many existing feature selection methods fall short in effectively leveraging time-series information, primarily because they are designed for static data. Our approach addresses this limitation by enabling the selection of time-varying feature subsets for each patient. Specifically, we employ reinforcement learning to optimize a policy under maximum cost restrictions. The prediction model is subsequently updated using synthetic data generated by trained policy. Our method can seamlessly integrate with non-differentiable prediction models. We conducted experiments on a sizable clinical dataset encompassing regression and classification tasks. The results demonstrate that our approach outperforms strong feature selection baselines, particularly when subjected to stringent cost limitations. Code will be released once paper is accepted.

}

\maketitle

\section{Introduction}

Predictive monitoring plays a crucial role in timely detecting acute illnesses, particularly in settings like the Intensive Care Unit (ICU), where clinical features such as lab tests and blood gas results are routinely collected based on doctors' requirements. These features may undergo multiple updates to track disease progression\cite{mimiciv, hirid, eicu}. While traditional predictive monitoring methods rely on basic vital signs like heart rate or respiratory rate to automatically trigger early alarms, they often fall short in providing further assistance.
 
Machine learning-based predictive monitoring, on the other hand, harnesses the power of multiple clinical features, offering more objective and accurate results than manual judgments. However, a significant hurdle in adapting machine learning models to real-world applications lies in feature acquisition. These models typically require a large volume of observations with high frequency to deliver better predictions\cite{hirid}, features readily available in extensive medical datasets like MIMIC\cite{mimiciv}. However, in reality, there exists a distribution shift in available features and their frequencies. For instance, frequent blood gas tests may lead to acquired anemia, which is often undesirable. Conversely, excessive efforts may be wasted in collecting unnecessary data, while critical signals might not be sampled adequately. To strike a balance between high prediction accuracy and low feature costs, it becomes imperative to determine which features should be collected and at what sampling frequency.

In this paper, we propose a novel method that leveraging reinforcement learning to navigate this delicate balance in both the feature space and the time axis. Initially, we train a predictor model using multi-variable time-series data. The predictor can be a neural network or a non-differentiable model(e.g. Decision Tree\cite{decision_tree}). Then we construct an environment based on the predictor and the training dataset. During each episode, sequences are randomly sampled from the training set, and the agent select a subset of features to update at each step. The feedback from the environment is twofold, consisting of cost rewards and prediction rewards. Once the policy converges, we create a new training set by recording how features are updated based on the agent's actions. Finally, a new predictor is trained on this generated dataset to align with new data distribution.

Our contributions can be summarized as follows:

1. We have introduced a novel method to tackle the feature selection problem within multivariate time-series scenarios, showcasing its superiority over existing methods. Our approach is adaptable to non-differentiable predictors and is applicable to both regression and classification tasks.

2. Additionally, we have presented an interpretation method to compute time-varying feature importance, offering more comprehensive insights along the time axis compared to existing methodologies. This not only enhances bedside monitoring in healthcare settings but also informs data collection strategies for clinical researchers.

In the following sections, we introduce related works in \autoref{sec:related_works}. In \autoref{sec:problem_definition}, we introduce the formulation of the time-series feature selection problem as a Partially Observable Markov Decision Process (POMDP) and explain the state transition process. We provide detailed explanation of our method in \autoref{sec:method}. In \autoref{subsec:reward}, we outline the novel reward design. The experimental setup and results are presented in \autoref{sec:exp}, followed by a thorough discussion on our method's performance and limitations in \autoref{sec:discussion}. Finally, we summarize our study in \autoref{sec:conclusion}.

\section{Related works} \label{sec:related_works}

Feature selection methodologies have been extensively explored in recent works\cite{chan14,nanfeng2017,Meng2018,ling2004,cron10,fahy19}, categorized into three main types: Filter methods, wrapper methods, and embedded methods.

Filter-based methods\cite{xu2013,Penar2010,ZhouQif2016, relief} evaluate individual features by their relevance or mutual information concerning the target variable. It works independently without the need of pre-trained models. Feature importance is computed by pre-defined metrics. Relief-based methods\cite{relief} operate by estimating the relevance of features based on their ability to discriminate between instances. This discrimination is computed by assessing the proximity between instances, thereby evaluating how changes in a feature impact the class label or target variable. Wrapper methods\cite{Jishihao2007, lvm} use the final objective functions to select important features. It directly regards the performance of trained model as its metric. Therefore it not only considers the feature relevance but also how they can improve the model. Las Vegas Wrapper\cite{lvm} evaluates subsets of features based on their performances. Only the most promising feature subset is selected. In Embedded methods\cite{ling2004,kusner2014, lasso}, model training and feature selection are combined to a unified process. LASSO\cite{lasso} is a powerful feature selection technique belonging to the realm of embedded methods. LASSO works by adding a penalty term (L1 norm) to the absolute magnitude of the coefficients in a regression model. 

However, there are few works focusing on multivariate time-series feature selection. In this domain, \cite{cron10} detects important segments in univariate time-series inputs. This method is applied on sensor observations or marketing data, which have significantly higher sample frequencies than clinical data. \cite{walg2006} focus on multiple time-series data, employing a cost-benefit model to estimate the optimal number of features for prediction, notably applied in forecasting energy production with sensor data.However, these approaches do not address the reduction of data frequencies and time-varying selection, as sensors can continuously provide data once installed. In clinical prediction, it is imperative to minimize the total number of tests due to the associated costs.

While static feature selection methods can also be applied in time-series feature selection. SHAP\cite{lund17} is widely applied for feature explanation of boosting trees, which plays an important role in time-series prediction. Although SHAP can not leverage time-varying data in computing feature importance, it provides reliable estimations by simply averaging along time ticks to compute importance for each feature.

\begin{figure}
    \centering
    \includegraphics[width=0.75\linewidth]{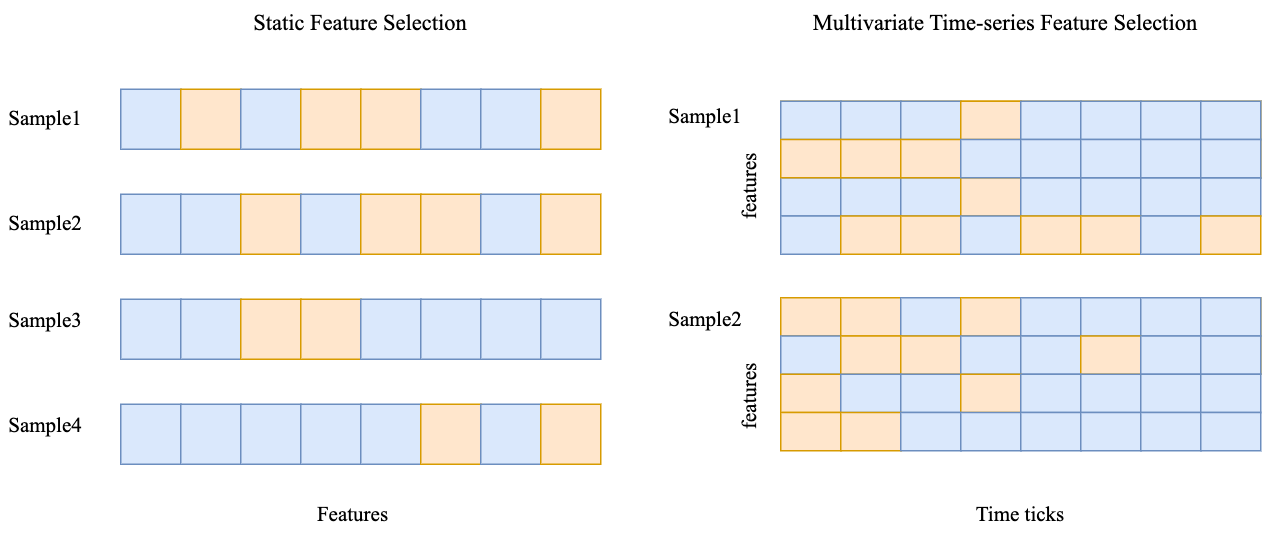}
    \caption{Different feature selection scenarios. Orange blocks indicate selected features. \textbf{Left}: Existed methods select a subset of features for each sample. In univariate time-series feature selection, horizon axis is the time ticks of a single feature. \textbf{Right}: Our method focus on finding time varying feature subsets for each sample sequence.}
    \label{fig:related_works}
\end{figure}

\cite{sataer2023, an2022, yu23} applied reinforcement learning with static clinical features. However, these methods do not consider the recurrence of the same feature in clinical sequences. The main difference between our method and existing feature selection methods is that we evaluate the additional information of features updated multiple times. We evaluate feature importance across features and their time variation, collecting time-varying feature subsets based on previous observations. For each patient and each time tick, the model output can be different. The differences are depicted in \autoref{fig:related_works}.

\section{Problem formulation} \label{sec:problem_definition}

In this section, we formulate generalized dynamic feature selection problem without specific approaches. We further analyze our method under reinforcement learning framework in \autoref{subsec:framework} .

We randomly partition all subject sequences into training set $D_{train}$, validation set $D_{val}$ and test set $D_{test}$. We use $(\bx^i,\by^i) \in D$ to denote the $i$-th data and label sequence. The sequence $\mathbf{x}^i$ is a 2D matrix in the shape of $(N_{F}, N_{T})$, where $N_F$ is number of features and $N_T$ is the total number of ticks. We use $\mathbf{x}^i_{j,k}$ to represent the value of $j$-th features in $k$-th ticks of $i$-th sequence. The label $\by^i$ is a vector in length $N_T$. A boolean matrix $\mathbf{a}^i$, in the same shape as $\mathbf{x}^i$, indicates whether each feature is updated at given ticks. $\mathbf{a}_{j,:}^i$ is usually not a one-hot vector since multiple features can be updated in one tick. 

Our model can only observe updated features. We denote observable states as $\mathbf{s}^i$.  We insert a initial state $\mathbf{s}^i_{:,0}$ as a constant vector to all sequences. For each episode, the sequence index $i$ is randomly sampled from dataset after initial state. The state transition is defined in \autoref{equ:state_transition}:

\begin{equation}
    \label{equ:state_transition}
    \mathbf{s}^i_{:,t}=\mathbf{s}^i_{:,t-1} \otimes \hat{\mathbf{a}}^i_{:, t} + \mathbf{x}^i_{:,t-1} \otimes \mathbf{a}^i_{:, t}
\end{equation}

Our model consists of a predictor $P_\phi$ and an actor $\pi_{\theta}$ with trainable parameters $\phi$ and $\theta$. We get $t$-th label prediction $p^i_t$ by current and historical states $p^i_t=P_\phi(\bs^i_{:,0:t+1})$ . The action is generated from $\ba^i_{:, t}=\pi_\theta(\bs^i_{:,0:t+1},\ba^i_{:,0:t})$. To evaluate cost and performance, we define a cost function $C(\ba^i, \bx^i)$ to calculate a scalar cost $c_i$ for each sequence. A loss function $L(\by, \bp)$ is used to calculate prediction loss. In regression tasks,  we use $L(\by^i, \bp^i)$ to calculate loss in each sequence. In classification tasks, it takes all sequences as input to calculate AUROC.

The goal is minimizing prediction loss with given maximum cost $C_{max}$ as shown in \autoref{eqn:goal} . All evaluations are performed in test dataset. 

\begin{align}
    \label{eqn:goal}
    \theta^*, \phi^* &=\arg\min_{\theta,\phi} L(\by, \bp;\pi_\theta, P_\phi)) \; s.t. C(\ba,\bx; \pi_\theta) \leq C_{max}
\end{align}

\begin{figure}
    \centering
    \includegraphics[width=0.5\linewidth]{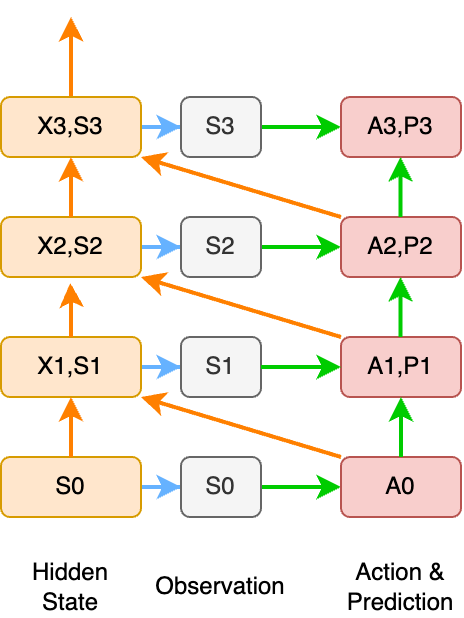}
    \caption{POMDP state transition. Orange arrows denote fixed conditional state transitions. Blue arrows represent a simple copy operation. Green arrows represent trainable functions. The initial state $s_0$ is a non-trainable constant vector. Sequence $\bx^i$ is randomly sampled from dataset.}
    \label{fig:pomdp}
\end{figure}

This problem has various connections with existing fields. Firstly, it is a Partial Observable Markov Decision Process(POMDP) if we combine $\mathbf{z}:=[\bs, \bx]$ as hidden state and view $\bs$ as observable. The POMDP state transition is depicted in \autoref{fig:pomdp}. Secondly, states $\bs^i$ are synthesized based on $\bx^i$ and $\ba^i$ . The data collection policy of $\bx^i$ is unknown and fixed. Therefore the training procedure has to be offline, which means the policy is running on a given dataset. Thirdly, in practice it is common to use cost regulation penalty instead of max cost limitation. It can be interpreted as reinforcement learning with multiple objectives.

\section{Methods} \label{sec:method}

In this section, we first distinguish this scenario with common reinforcement learning scenarios in \autoref{subsec:framework}.  Training algorithm and basic implementations are described in \autoref{subsec:training}. We describe reward design and predictor update in the following subsections \autoref{subsec:reward} and \autoref{subsec:predictor}.

\subsection{Reinforcement learning framework} \label{subsec:framework}

The goal within our framework is maximize accumulate reward exception $\mathbb{E}_{\bx, \by \in D_{train}}[\sum^T_{i=0} \gamma^i R(\bs, \bx, \by)]$. This aligns with the objective in \autoref{eqn:goal} under several assumptions: 1. The distribution shift between $D_{train}$ and $D_{test}$ is negligible. 2. The discount factor $\gamma$ is sufficiently close to 1. 3. Actor and critic are near convergence when the cost reward is decreasing. Assumption 1 can be satisfied when dataset is large, as the training and test sets are i.i.d. sampled. We found that approximately 5000 samples are enough based on our experiments. For assumption 2, the effect of $\gamma$ depends on tasks. Estimating the state value is challenging due to varied sequence lengths, and credit assignment becomes more difficult when $\gamma$ is near 1. We use $\gamma=0.8$ or $\gamma=0.95$ in our experiments. Assumption 3 arises from the penalty method. The policy should be close to optimized at each step when the actual cost is decreasing. Therefore, convergence to an optimal policy with cost restriction is guaranteed. However, it's important that assumption 3 may be violated with linear cost rewards. When the cost coefficient is large, the initial cost decreases rapidly without sufficient exploration. We further discuss this issue in \autoref{subsec:reward}.

\begin{algorithm}
\caption{Training}\label{alg:full_alg}
\begin{algorithmic}
\State Initialization: Dataset $D_{train}$, $D_{valid}$, $D_{test}$, Actor $\pi_\theta$, Critic $V_\psi$, Predictor $P_\phi$
\State $C_{valid} = +\infty$
\State Train predictor $\phi \gets \phi_0$ with dataset $(\bx^i, \by^i) \in D_{train}$
\While{$C_{valid} > C_{max}$}
    \Comment{Rollout}
    \For{batch=1,2,... }
       \State Sample a batch $M \subset D_{train}$
       \State Compute $\ba^i, \bs^i$ for every $(\bx^i, \by^i) \in M$
    \EndFor
    \State  Compute $R_{pred}$ by \autoref{equ:pred_reward}
    \State $G_{cost} \gets \frac{1}{1+\exp{[\alpha (1-C_{train}/C_{max})]}}$
    \State Compute $R_{cost}$ by \autoref{equ:reward_cost}
    \State $\mathbf{r} \gets R_{pred} + R_{cost}$
    \Comment{Policy Update}
    \State Update $\theta, \psi$ using buffer $(\mathbf{r}, \ba, \bs)$
   \State  Evaluate $C^{iter}_{valid}$ on $D_{valid}$
    \If{ $C^{iter-1}_{valid} - C^{iter}_{valid} > \Delta_{thres}$}
        \State $\beta \gets \min(1.5\beta, \beta + \Delta_{\beta})$
    \EndIf
\EndWhile
\State Generate $\bs^i$ for every  $(\bx^i, \by^i) \in D_{train}$, $D_{train} \gets D_{train} \cup \{ \bs^i\}$
\State Train predictor $\phi \gets \phi^*$ with  $(\bs^i, \by^i) \in D_{train}$
\end{algorithmic}
\end{algorithm}

\subsection{Training procedure} \label{subsec:training}

The training algorithm is described in \autoref{alg:full_alg}. Given dataset $D_{train}$, $D_{val}$ and  $D_{test}$, we first train a predictor $P_\phi$ to predict sequential labels with supervision. There are multiple choices of prediction model. We implement $P_\phi$ as a Gradient Boosted Decision Tree(GBDT)\cite{catboost} or LSTM\cite{lstm}. For GBDT, the input are flattened and we only use the current features $\bx^i_{:,t}$ to predict corresponding label $\by^i_{:,t}$. Historical features are used in LSTM predictor. There is no cost limitation in training predictors. 

Then we freeze the trained predictor to optimize policy. We use a widely adopted Actor-Critic algorithm PPO\cite{ppo}. Both actor and critic are implemented as LSTM with observation input $\bs^i_{:,0:t+1}$ in tick $t$. In each step, the actor $\pi_\theta(\bs^i_{:,0:t+1},\ba^i_{:,0:t})$ outputs logits in the shape of $(N_F, 2)$. They form $N_F$ Bernoulli distributions. In $\ba^i_{:,t}$ , each action is independently sampled from a distribution. New observation $\bs^i_{:,t+1}$ is generated according to \autoref{equ:state_transition}.  When $\bs^i$ is generated, label prediction is calculated by $\bp^i =P_\phi(s^i)$. Once the rollout phase is finished, cost rewards and prediction rewards are generated for each action vector $\ba^i_{:, t}$. We further explain the reward design in \autoref{subsec:reward}. Policy is trained on $D_{train}$ and stopped when the cost is under $C_{max}$ in $D_{val}$. For more implementation details, please refer to \autoref{subsec:imp_details}. 

Noticed that $P_\phi$ uses $\bx^i$ in predictor training but $\bs^i$ in policy training. It will lead to a distribution shift if $P_\phi$ is not updated. Although predictor implemented as a neural network can be updated concurrently\cite{yu23}, it is relatively difficult to jointly optimize $P_\phi$ and $\pi_{\theta}$ when $P_\phi$ is non-differentiable. We perform an iterative update as \autoref{eqn:iter_update}. We denote the predictor trained with $\bx^i$ as $P_{\phi_0}$. When policy is converged, we update $P_\phi$ using synthesized data $\bs^i$. We further explain the predictor training in \autoref{subsec:predictor}.

\begin{align}
    \label{eqn:iter_update}
    \theta^* &=\arg\min_\theta L(\by, \bp;\pi_\theta, P_{\phi_0})) \; s.t. C(\ba,\bx; \pi_\theta) \leq C_{max} \\
    \phi^* &= \arg \min_\phi L(\by, \bp;\pi_\theta^*, P_\phi)
\end{align}

At inference phase, we calculate the average cost $c_{test} = \frac{1}{N_{test}} \sum_{i=1}^{N_{test}}C(\ba^i, \bx^i)$ and model performance $L(\by, \bp;\pi_{\theta^*}, P_{\phi^*})$ on test set $D_{test}$. The form of cost and loss functions depend on tasks. Please refer to \autoref{sec:exp} for further details.

\begin{figure}
    \centering
    \includegraphics[width=1.0\linewidth]{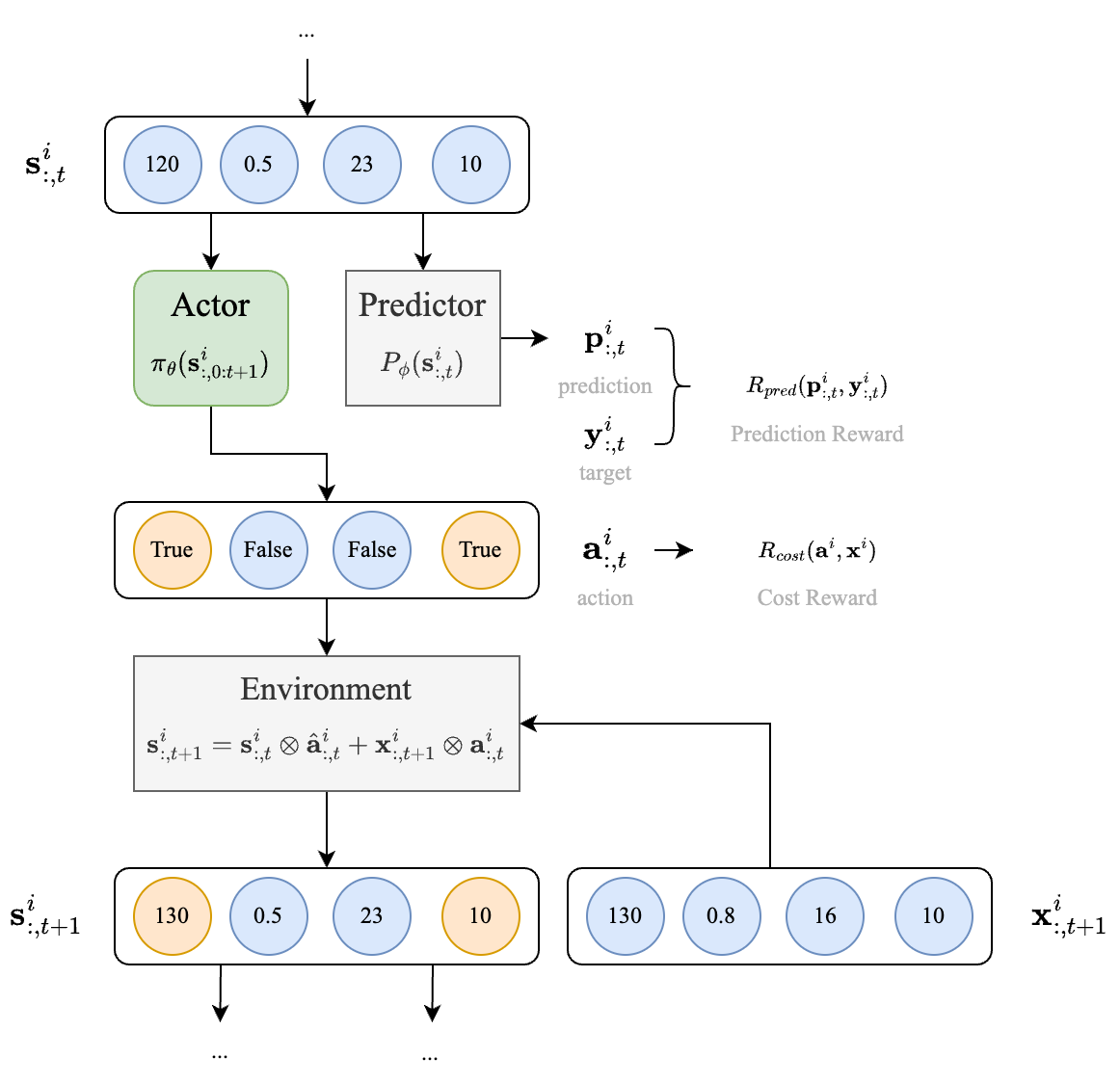}
    \caption{How actor cooperates with predictor in synthetic environment.  Orange circles: Updated features in next tick. Blue circles: Features that not updated. Grey blocks: environment and predictor are fixed in policy training.}
    \label{fig:framework}
\end{figure}

\subsection{Reward design} \label{subsec:reward}

The reward in each step consists of two parts: cost reward and prediction reward. The main challenge of reward design is to keep a balance between these two parts. All rewards are computed for each tick in a sequence.

\subsubsection{Prediction reward}

In regression tasks, we use Mean Absolute Error as the loss function. Hence it is natural to use MAE as reward function. However, we found that it is hard for critic to estimate which state will lead to bad estimation. These samples have large MAE loss, which lead to unstable advantage values. Our approach is to normalize MAE loss based on the pre-trained predictor $P_{\phi_0}$. We calculate MAE loss from $P_{\phi_0}$ as $L_{baseline}$. Then normalize $L_{MAE}$ to calculate the relative performance. $L_{eps}$ is used to prevent too small denominator.
\begin{equation}
    \label{}
    \hat R_{pred}^{reg}=\frac{L^{MAE}_{pred}-L^{MAE}_{baseline}}{\mathrm{max}\left(L^{MAE}_{baseline},L_{eps}\right)}
\end{equation}

In classification tasks, we adapt a pairwise loss function to optimize AUROC\cite{pairauc}. For each batch in rollout phase, we randomly generate pairs for each prediction. Two predictions in a pair have different labels. Then we apply a pairwise loss for each prediction. In \autoref{eqn:pred_cls}, $M$ contains all predictions in a rollout batch. $y^i$ and $y^j$ are scalar predictions in tick level. Label values are in $[-1,1]$.
\begin{equation}
    \label{eqn:pred_cls}
    \hat R_{pred}^{cls}=y^i(p^i-p^j) , \; j \in \{j|j\in M, y^j=-y^i \}
\end{equation}

We found that the scale of prediction reward may change in different tasks. We add a reward scaling to output a unified scale of prediction reward. In policy training, we do not use additional reward scaling or advantage normalization.

 \begin{equation}
    \label{equ:pred_reward}
     R_{pred} = \frac{\hat R_{pred}}{\frac{1}{N}\sum^N_0|\hat R_{pred}|},
     \hat R_{pred} = \left\{ \begin{aligned} 
        \hat R_{pred}^{reg} & \text{  for regression} \\
        \hat R_{pred}^{cls} & \text{  for classification}
     \end{aligned} \right.
 \end{equation}
 
\subsubsection{Cost reward}

A naive per-step cost reward is the negative proportion of feature updates. Actions resulting in fewer feature acquisitions receive less punishment. Existing methods\cite{an2022, yu23} often employ a linear cost reward, which multiplies a constant cost coefficient by the proportion of feature updates. However, this approach presents two drawbacks: 1. Controlling the final cost becomes challenging as the prediction reward may impede cost reduction at an unknown point. 2. When $C_{max}$ is small, excessively strong cost rewards can lead to suboptimal convergence because there is limited exploration during the initial cost reduction process.

Our approach involves dynamically controlling the cost reward coefficient during the training process. We initially apply a gate multiplier to the linear proportion cost reward (\autoref{equ:gate_coeff}). An exponentially smoothing cost $C_{train}$ tracks current training cost. Once it satisfies max cost limitation, typically lower than initial cost, the gate multiplier will rapidly decrease to zero. Consequently, the final cost remains the desired range. The smoothing parameter $\alpha$ ensures the stability of this process. When $C_{max}$ is small and $C_{train}$ is close to $C_{max}$, the total cost reward will be near zero even when $G_{cost}$ is near 1. We add a constant reward $C_{base}$ to give enough punishment in this situation.

\begin{equation}
    \label{equ:gate_coeff}
    G_{cost} = \frac{1}{1+\exp{[\alpha (1-C_{train}/C_{max})]}}
\end{equation}

To address the second problem, we adopt a strategy of setting a small initial cost coefficient $\beta$ and gradually increasing it when the validation cost shows no decrement. This approach is similar to the reduce on plateau technique in adjusting learning rate. In practice, we monitor the last three validation costs. If the rate of cost decrease falls below a predefined threshold, we increment the cost coefficient by a constant factor $\Delta_\beta$ . We use a smoothed update function $\beta \gets \min(1.5\beta, \beta + \Delta_{\beta})$ to avoid too strong punishment when $C_{max}$ is relatively large.

Finally, the cost reward used in our method can be written as \autoref{equ:reward_cost}. Features have different costs $c^{dyn}_k$ or $c^{sta}_k$ related to their type, per-tick cost and whether it is the first fetch. The per-tick cost is defined in two cost settings(\autoref{subsec:metrics}). 

\begin{align}
    \label{equ:reward_cost}
    R_{cost} &= G_{cost}(\beta \frac{1}{N_{fea}} \sum^{N_{fea}}_{k=1}a^i_{k,t}c^i_{k,t} + C_{base}) \\
    c^i_{k,t} &= \left\{ \begin{aligned} 
        & c^{dyn}_k & \text{  if k-th feature is dynamic} \\
        & c^{sta}_k(1-\max(\ba^i_{k,:t})) & \text{  if k-th feature is static}
     \end{aligned} \right.
\end{align}

\subsection{Predictor update} \label{subsec:predictor}

In time-series prediction, Gradient Boosted Decision Trees (GBDT) are indispensable compared to Neural Networks. We aim for our method to be generalizable to non-differentiable models, enabling the predictor to even be an ensemble model. This goal is achieved by iteratively training the predictor and policy. While multiple iterations may further enhance performance, we have observed that one additional update is typically sufficient to train a good predictor.

When predictor is a GBDT, we use $p^i_t=P_\phi(\bs^i_{:,t})$, indicating that only the current state is used for prediction. For LSTM predictor, historical data can be utilized as $p^i_t=P_\phi(\bs^i_{:,0:t+1})$. During the first predictor training, the predictor input is the normalized $\bx^i$ concatenated with initial state $\bs_{:, 0}$. The initial state is a non-trainable fixed vector. The pre-trained predictor $P_{\phi_0}$ often uses more features than final predictor. When the policy has converged, some features may not be updated throughout the sequence, but the actor cannot block these features in the predictor. We generate $\bs^i$ in training set based on state transition( \autoref{equ:state_transition}) and current policy. Since $\bs^i$ always satisfies $C_{max}$ limitation, we train a new predictor to predict $\by^i$ without cost restriction.  

We have found that final performance largely depends on the initial policy behavior. It is more challenging to identify valid features when the initial cost is much smaller than $C_{max}$. We set initial sampling probability close to 1 (i.e. 0.8) so that $\bs^i$ closely resembles $\bx^i$ at the start of training.  When $C_{max}$ is very large, our experiments show that there is no performance degradation compared to directly use the pre-trained $P_{\phi_0}$.

\section{Experiments}  \label{sec:exp}

\subsection{Evaluation metrics} \label{subsec:metrics}

To evaluate the overall performance of our method, we draw a cost-loss curve by training the model with a range of \(C_{\text{max}}\) values and observing the real costs in the test dataset. A better method has a curve that lies below others. All features in the dataset are divided into static features (e.g., age), which have only one recorded value in a sequence, and dynamic features, where each update incurs a cost. For each task, we design two types of cost functions: The first cost function, \texttt{simple\_cost}, adds a unit cost when a static feature is updated for the first time or when a dynamic feature is updated. We assert that each bio-test in reality incurs a cost even if it has the same value as the last observation. The second cost function, \texttt{complex\_cost}, assigns different costs to different features. Since each feature has different sampling frequencies, we first estimate the cost of one observation for features based on their categories. Then we compute the equivalent cost per tick for all features. An expensive feature may have a low per-tick cost if it rarely occurs in a sequence. In \texttt{complex\_cost}, if a dynamic feature is updated, it adds a per-tick cost to the total cost. Static features have a per-tick cost equal to the cost of one observation because they are updated no more than once in a sequence.

\subsection{Baseline methods}

As far as we know, there are no similar methods focusing on dynamic time-series feature selection. To establish a baseline, we adapt three methods to compute feature importance: the widely used feature selection method SHAP\cite{shap}, LASSO\cite{lasso} and Support Vector Machine(SVM) with L1 penalty. In experiments using the \texttt{simple\_cost} setting, we select the top-k important features to satisfy $C_{max}$ limitations. Baseline predictors are then trained using a subset of features with different costs. In the \texttt{complex\_cost} setting, an average sequential cost is computed for each feature. To find the optimal feature subset, we employ dynamic programming to select a subset of features that maximize the sum of feature importance under a given sequential cost limitation. The subset is fixed across subjects given a max cost limitation. We discuss this setting in  \autoref{subsec:results}. For fair comparison, we apply the same cost evaluation criterion for our method and baseline methods. 

We use GBDT and LSTM as predictors in the two tasks. For the baseline methods, we first run a GBDT to compute SHAP values and then train a predictor on the given feature subsets. The predictor can be either a GBDT or an LSTM. The feature importance computed by LASSO is used to train a linear regression model. We train a logistic regression model with feature importance from SVM with L1 penalty in classification tasks.

In summary, for each experiment and cost setting, we run three baseline methods.

\begin{enumerate}
    \item GBDT: We use GBDT for both computing SHAP importance and label prediction. 
    \item LSTM: We use GBDT for computing SHAP importance and LSTM for label prediction.
    \item LASSO: In regression task, we use LASSO for computing importance and linear regression for label prediction.
    \item SVM: In classification task, we use SVM-l1 for computing importance and logistic regression for label prediction
\end{enumerate}

\subsection{Dataset and Task description} \label{subsec:data_and_task}

We evaluate the performance of our method using a large clinical dataset, MIMIC-IV\cite{mimiciv}. While other datasets such as HiRID\cite{hirid} and eICU-CRD\cite{eicu} also contain time-series clinical data, MIMIC-IV stands out due to its adequate sampling frequency of blood gas tests, which are more cost-sensitive than vital signs.

We have designed two time-series prediction tasks: P/F ratio prediction and ventilation termination prediction. The first task is a regression task, and its performance is evaluated using the Mean Absolute Error (MAE). The second task is a binary classification task, and we use the Area Under the Receiver Operating Characteristic curve (AUROC) as the evaluation metric.

\subsubsection{P/F ratio prediction}

The P/F ratio serves as a crucial indicator for Acute Respiratory Distress Syndrome (ARDS), a life-threatening condition, particularly for patients with severe illnesses. It is calculated by dividing the Pressure of Arterial Oxygen (PaO2) by the Fractional Inspired Oxygen (FiO2). The possible range of P/F ratio is $[0, 500]$, with values below 300 considered dangerous. To predict the future value of P/F ratio, we build a regression model.

We collected data from 8126 patients with sepsis, indicating severe illness, extracted using MIMIC-Code\cite{mimic-code}. Data cleansing involved removing patients with excessively long ICU stays ($> 96$ hours), features with high missing rates ($> 50\%$), and patients without P/F labels. Using half-hour increments as unit tick length, we noted an average ICU stay length of 49.2 hours in the training set. To prevent data leakage, we removed PaO2 and FiO2 from the dataset, as they are target-related features. For each time tick, the label represents the minimum P/F value in the subsequent 8-hour prediction window.

\subsubsection{Ventilation termination prediction}

Ventilation termination represents a critical decision in clinical settings, with significant impacts on patient outcomes and safety. Premature termination can lead to respiratory failure, while prolonged ventilation support increases the risk of infection and extends ICU stays. To predict when to stop ventilation, we develop a binary classification model.

Our dataset comprises a total of 12,802 patients who have received ventilation support. We extract cohorts based on MIMIC-Code and apply the same data cleansing criteria as for P/F ratio prediction. For each time tick, the label is set to positive if the current ventilation status is positive and there exists a negative status in the subsequent 4 hours, indicating an upcoming ventilation termination. Only a short period before termination will have positive labels. The label distribution is highly unbalanced, with only 9.5\% positive labels. In predictor and baseline training, we adjust label weights according to the opposite label proportion: $[W^n, W^p] = [\frac{N^p}{N^p+N^n}, \frac{N^n}{N^p+N^n}]$. 

\subsection{Results} \label{subsec:results}

\begin{figure}
    \centering
    \includegraphics[width=1.0\linewidth]{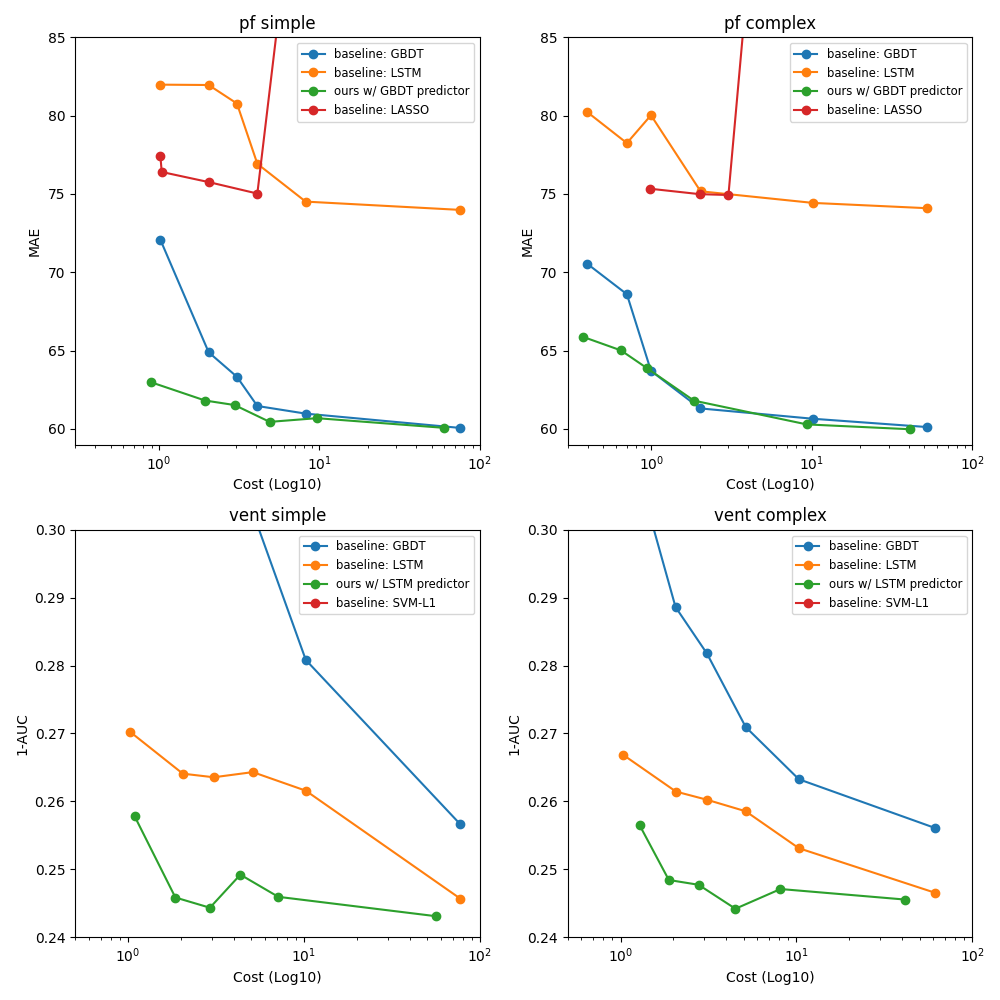}
    \caption{Performance comparison results on test dataset. We run four baseline methods (GBDT, LSTM, LASSO, SVM-L1)  for comparison. For our method, we use GBDT as predictor in P/F ratio prediction and LSTM in ventilation prediction. The cost is computed as average per-tick cost. We set X axis as logarithmic based on 10. For regression task, we use MAE as loss function. We use 1-AUC as loss function in the binary classification task. Some curves of LASSO and SVM-L1 are not drawn entirely. The detailed results are provided in \autoref{subsec: eval-results}.  }
    \label{fig:performance}
\end{figure}

The main results are shown in \autoref{fig:performance}. In P/F ratio prediction, the GBDT baseline achieves an MAE loss of 60.06 without cost limitation ($C_{\text{max}}=100$). Other two baseline methods have significantly worse loss curves. Our method yields similar results ($L^{\text{MAE}}=60.07, C_{\text{max}}=100$), indicating that our method does not cause a performance degradation when there is no cost restriction. The trained cost of our method is slightly lower than the initial cost (59.42 vs 74.88). This slight difference may be attributed to the initial bias $p=0.8$ (see \autoref{subsec:imp_details}). Our method demonstrates superiority when the cost becomes relatively low. In P/F ratio prediction, the performance gap becomes significant when the target cost is roughly under 5\% of the original. In the ventilation task, this threshold is about 10\%. In an extreme case, our method maintains an MAE loss of 62.98 with a cost of 0.89, significantly lower than the GBDT baseline (MAE loss of 72.08, cost of 1.02). Similar trends were observed in other experiments, with our method exhibiting a relative performance decrease of no more than 8\%, indicating that the models can still be effective under low cost restrictions.

In the ventilation task, our method achieves the same performance as the LSTM baseline when there is no cost limitation ($L^{\text{AUC}}=0.243$ in our method, $L^{\text{AUC}}=0.243$ in the LSTM baseline). Other two baselines perform worse in all cost limitations. When there is no cost restriction, the objective in policy update shifts to maximizing AUC rather than minimizing cross-entropy loss, which is used in prediction training. This indicates that the change in objective does not affect performance when using label weights in cross-entropy loss. However, when the cost is extremely low, our method achieves better performances but also experiences a performance degradation. This may be because the ventilation task requires more features than the P/F prediction task, as observed in the following result visualization.

Our method can also serve as an interpretation method. Specifically, we compute the expectation of activation for each Bernoulli distribution. For an action of fetching feature $i$ in tick $t$, its activation value is computed by \autoref{equ:visualize}. Action sequences are truncated to a max length $t \leq T_{max}$. 

\begin{equation}
    \label{equ:visualize}
    \hat \ba_{i;t}(c)=\mathbb{E}_{\bx^i \in D_{test}}[\mathbb{I}_{\ba^i_{i;t}>0} | C_{max}=c, \pi_\theta, P_\phi]
\end{equation}

\begin{figure}
    \centering
    \includegraphics[width=1\linewidth]{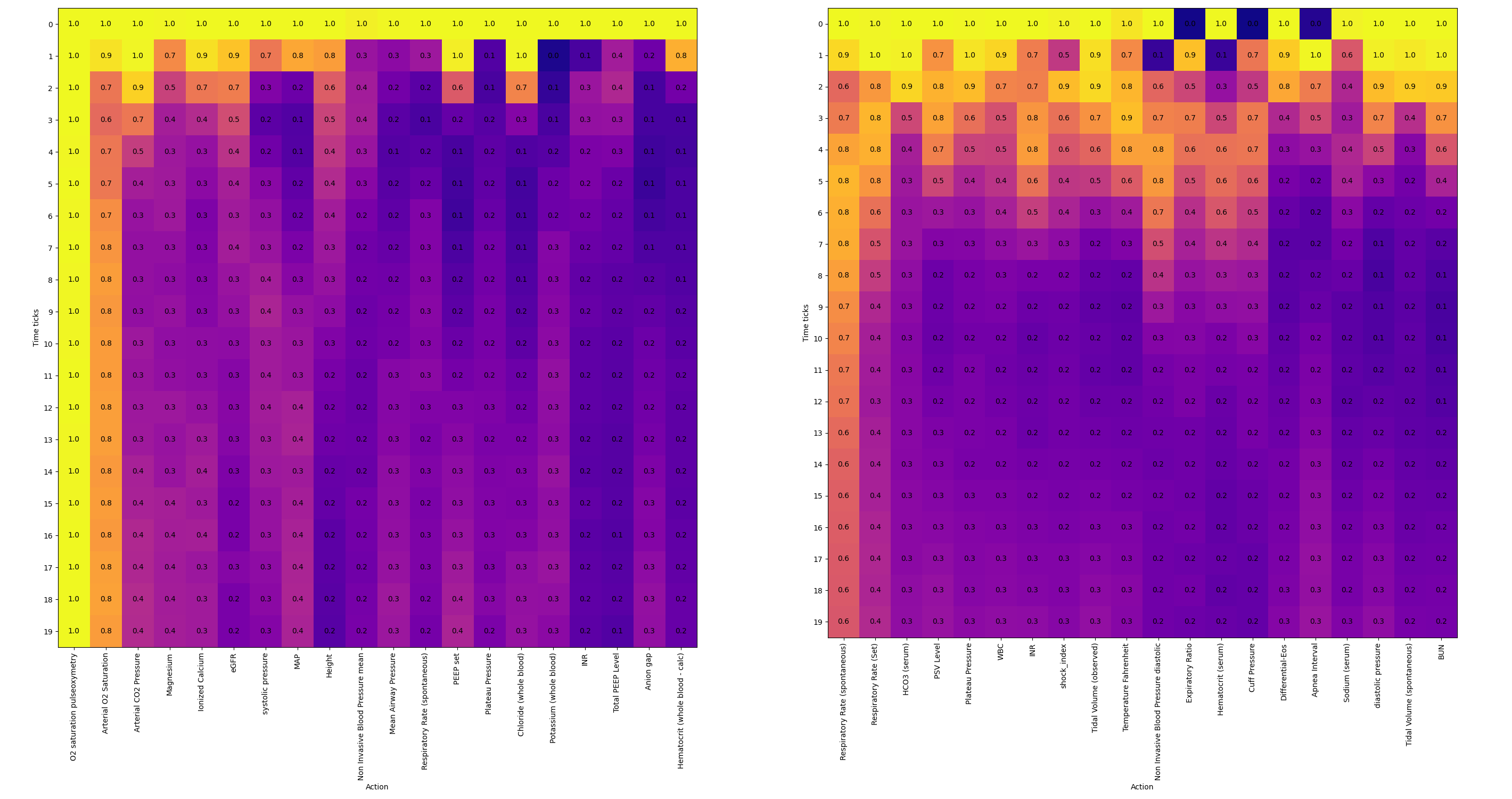}
    \caption{Policy visualization in two tasks. X-axis represents ordered features, and Y-axis represents time ticks (tick=0 at the top). We use $C_{\text{max}}=10$ and simple cost setting in this figure. Features are ordered by the mean activation along the time axis. \textbf{Left:} In P/F prediction task, feature selection is more concentrated on specific features. Some features are sampled multiple times while others are only sampled in the initial state. \textbf{Right:} In ventilation task, the feature selection is dispersed.}
    \label{fig:vis}
\end{figure}

The activation distribution is depicted in \autoref{fig:vis}. We found significant feature preferences both along the feature axis and the time axis. These preferences are influenced by the intrinsic properties of the task. In both tasks, a few important features exhibit high sample frequency even when cost limitations are restrictive. In the P/F prediction task, O2 saturation pulseoxymetry and Arterial O2 saturation have more than 80\% activation in the first 20 ticks when $C_{\text{max}}=10$. In comparison, the activation of other features rapidly decreases to below 40\% after a few ticks. In the ventilation task, Respiratory Rate (spontaneous) exhibits higher activation, but the activation distribution is more dispersed. There are more features sampled multiple times in the ventilation task, indicating that timely collection of the latest features can help improve prediction performance. 

Our method tends to apply a similar selection policy across different subjects, suggesting that selecting the same feature subsets in baseline methods is not the reason for the performance gap compared to our method. However, our policy applies different selection frequencies along the time axis, a characteristic that cannot be replicated by static feature selection methods.

An interesting phenomenon observed is that some features are sampled in the first tick but not updated until several ticks later. This could be because continuously sampling a feature provides no additional information until a new value is provided. Additionally, we observed that a large set of features provides useful information only in their initial values.

\subsection{Ablation study}

\begin{figure}
     \centering
     \includegraphics[width=1\linewidth]{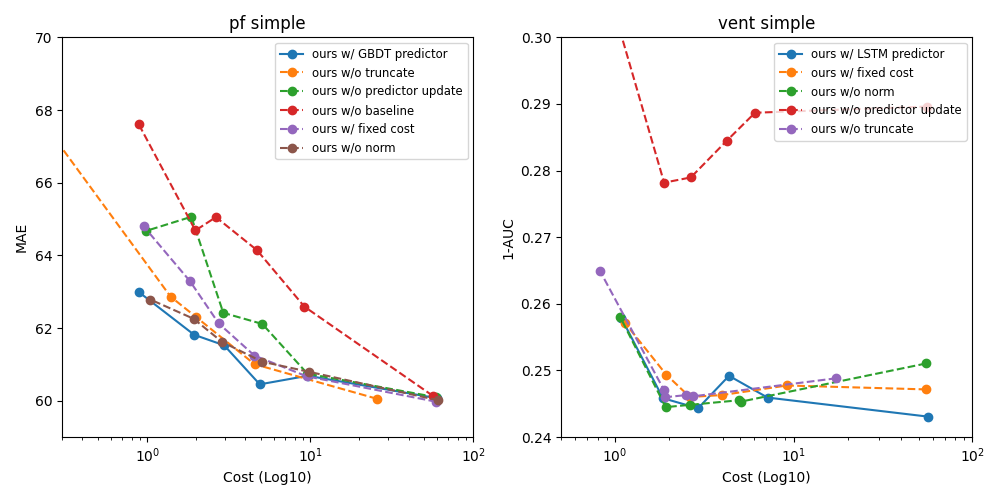}
     \caption{Ablation result. Dashed lines are ablation studies while the solid line is our model's loss curve with no ablation. \textbf{Left}: results in P/F ratio prediction task. \textbf{Right}: results in Ventilation task. X and Y axis are the same as \autoref{fig:performance}.  }
     \label{fig:ablation}
 \end{figure}

We conduct five ablation settings. The results are depicted in \autoref{fig:ablation}. Ablation studies are conducted on the \texttt{simple\_cost} setting.

\textbf{No Predictor Update}: In this setting, we remove the second predictor update to observe if data distribution shift causes performance degradation. The results in both tasks show significant performance degradation without the second predictor update, highlighting the importance of synchronizing predictor updates with policy updates.

\textbf{No Baseline}: We remove baselines in computing the prediction reward and use $\hat{R}_{pred} = - L^{MAE}$ as the prediction reward instead. This ablation is performed only in the P/F prediction task because baselines are not applied in the ventilation task. The results indicate a performance degradation, suggesting that using baselines in computing prediction rewards is beneficial.

\textbf{Fixed Cost Coefficient}: We replace the dynamic cost reward with a fixed cost coefficient, effectively setting $\Delta_\beta = 0$ and $C_{base} = 0$ in \autoref{alg:full_alg}. The initial $\beta$ varies with different $C_{\text{max}}$ to achieve the actual cost. The gate multiplier $G_{\text{cost}}$ is kept enabled to obtain a controllable final result. In the P/F prediction task, we observed increased loss with low cost, probably due to too strong punishment in the initial cost reduction process. In the ventilation task, we observed slight performance degradation with high cost, possibly due to $C_{base} = 0$.

\textbf{No Truncate}: In this setting, we remove $G_{\text{cost}}$ but keep the dynamic $\beta$ update enabled. Using the gate multiplier allows the cost reward to quickly reduce to 0 when $C_{\text{max}}$ is satisfied, helping achieve a controllable actual cost. In experiment results, while the performance is retained, the trained cost becomes unstable. 

\textbf{No Normalization}: We also investigate the effect of prediction reward normalization mentioned in \autoref{equ:pred_reward}. Our observations indicate that keeping the relative scale of prediction reward and cost reward is necessary. Removing this normalization does not significantly affect the P/F prediction task. The right end of the loss curve in the ventilation task slightly increased, but other parts show no significant change. Thus, we keep prediction reward normalization enabled as it does not harm performance and is necessary for generalizing to other tasks.

\section{Discussion} \label{sec:discussion}

\textbf{Data distribution shift}: In our experiments, we encounter two potential distribution shifts. Firstly, optimizing the policy on the training set while testing its performance on the test set carries the risk of overfitting. However, we observe only a negligible performance decrease ($\leq 1\%$), likely because the credit assignment is not explicit enough to allow the model to memorize the training set. Secondly, there is a shift due to asynchronous training of the predictor and policy. Retraining the predictor multiple times is an option, but we found that retraining a GBDT predictor can lead to poorer performance when the policy is not converged. Once an important feature is ignored by the GBDT, it's difficult to keep it chosen by the optimized policy.

\textbf{POMDP}: We opt for Proximal Policy Optimization (PPO) as the training algorithm because it's a simple yet robust method across various tasks. Although PPO is designed for Markov Decision Processes (MDP), there have been successful approaches using PPO in Partially Observable Markov Decision Processes (POMDP), even with neural networks as function approximators. We believe other reinforcement learning algorithms could also be viable options for this study, and further research could explore differences in this aspect.

\textbf{Offline Training}: One fundamental difference between our approach and common reinforcement learning scenarios is that the policy cannot collect information in the real environment. We address this by using linear interpolation to mimic how vital signs change with no observation. We admit that this approximation may not be accurate enough due to limited sample frequency. For further study, a tailored sampling strategy can mitigate this problem to a degree.

\textbf{Feature cost}: Our experiments have shown that GBDT 's performance has an approximate linear relationship with the logarithm of cost when cost limitation is not very restrict. In reality, the actual feature cost may increase unexpectedly because there are many potential disease developments that need biology tests. It is possible to reduce the total feature cost by decreasing the test frequencies based on the time varying feature importance. Our interpretation method helps decide different test frequencies for each biology test. 

\section{Conclusion} \label{sec:conclusion}

In this paper, we proposed a novel method for dynamic time-series feature selection in clinical prediction monitoring scenarios. Our method leverages reinforcement learning to optimize feature subsets over time, allowing for improved prediction accuracy while considering cost constraints. We introduced a policy-based approach that dynamically selects features based on their relevance and timing, providing more accurate and interpretable results compared to existing feature selection methods. Through experiments on a large clinical dataset, we demonstrated the effectiveness of our method in both regression and classification tasks, outperforming strong feature selection baselines such as SHAP, particularly under strict cost limitations. Additionally, we conducted ablation studies to further validate the efficacy of our approach. 

There are also limitations in this work. Our method, based on reinforcement learning, has lower sample efficiency than traditional methods and takes more time to search a larger action space as the number of features increases. The offline training procedure may pose unexpected obstacles when applied in real-world interactive situations. Additionally, we acknowledge that interventions such as surgeries, medications, and other treatments are not fully investigated in our study, and the prediction results may be significantly influenced by them.

\section{DECLARATIONS}

\subsection{Availability of data and materials}

MIMIC-IV(version 2.2) is available at: \href{https://physionet.org/content/mimiciv/2.2/}{https://physionet.org/content/mimiciv/2.2/}

\subsection{Competing interests}

The authors declare that they have no competing interests.

\subsection{Funding}

This paper was funded by the National Key R\&D Program of China (No. 2021YFC2500803)

\subsection{Authors' contributions}

YC did the coding and writing.

\bibliography{main}

\section{Appendix}

\subsection{Implementation Details} \label{subsec:imp_details}

\textbf{Actor}: The actor is a LSTM. A dense layer is appended to LSTM output. All LSTMs in our experiments have 256 hidden states. The dense output dimension is the twice of feature number. We reshape the output into $(N_{batch}, N_{action},2)$. We apply softmax to the last dimension to get the probabilities of each Bernoulli distribution. We set initial selection probabilities to $0.8$ by adding a bias to output logits. The initial input state is filled with -4, which is alomost the lowest value in normalized feature distributions. Setting missing value to minimal is consistent with Catboost\cite{catboost} implementation. The initial hidden state is set to zeros. All input features are normalized by means and standards in training set. 

\textbf{Predictor}: There are two types of predictor. We use Catboost\cite{catboost} as GBDT predictor. The loss function is 'MultiClass' in both tasks and the evaluation metric is 'AUC' for classification and 'MAE' for regression. We training Catboost with iteration=800, max depth=5, learning rate=0.03. For each predictor we train 5 Catboost classifier or regressor with fixed seeds. the predictor output is the average of each models' output (probabilities for classification task). The LSTM predictor is a LSTM appended with a dense layer, which has output dimension of 1 for regression task and 2 for classification task. In predictor training, we use Adam optimizer and 1e-3 initial learning rate. We use a learning rate scheduler $\eta(p)=\max(10^{-4}, \eta_0(1-p^2))$ where p is the progress in $[0,1]$. We train the LSTM predictor for 20 epochs. It is enough since we observed LTSM will quickly overfit the training data. The best predictor is selected based on their performance in validation set.

\textbf{Dataset processing}: We randomly divide the whole dataset into 68\%, 12\%, 20\% for training, validation and test set. We removed features with more than 50\% missing rate(i.e. in more than 50\% subjects, we can not find at least one record of this feature). For P/F ratio prediction, we remove all admissions without target associated features(PaO2 and FiO2). Admissions less than 2 hours or longer than 96 hours are removed. We select patients who have sepsis roughly at the start time to ensure they have severe physical conditions. We use an interpolation step of 0.5 hour. Abnormal labels such as PaO2 $\geq$ 500 or FiO2 $\leq$ 21\% are clipped to the border values.

\textbf{PPO training}: We use Adam optimizer and learning rate=1e-3 for both actor and critic. In regression task, we set $\gamma=0.8$ and $\lambda=0.95$. We set $\gamma=0.95$ in classification task. The $\lambda$ is close to 1 to avoid value bias since critic can not get hidden state in POMDP setting.  We set PPO clip epsilon to 0.2. For each update, we collect 20000 ticks of data and update 5 times for 2 minibatches. We use gradient clip and set eps=1e-5 for Adam optimizer.  We set a minimal total step 5e6 and maximum step of 4e7. In P/F ratio task, the average step for a sequence is around 50. In ventilation task the average step is 42.

\textbf{Cost hyper-parameters}: We set $\beta=5$ in initial training and $\Delta_\beta=5$ for each increment. $C_{base}$ is set to 0.2. We update $\beta$ when the last three validation loss decrement is slower than 0.5 per 1e6 steps. These hyper-parameter setting is fixed in two tasks.

\textbf{Baseline hyper-parameters}:  For lasso baseline, we use Lasso with cross validation(LassoCV) with eps=0.001, n\_alphas=100, max\_iter=20000. Lasso converged under this max iteration. For SVM-L1, we train SVM Classifier with squared hinge loss, max\_iter=1000, C=1.0, tol=1e-4 and balanced class weight. After computing optimal feature subsets, we train Logistic Regression model with max\_iter=20000, tol=1e-4 and balanced class weight. All baseline methods are implemented by sklearn.

\subsection{Cost settings} \label{subsec:cost_set}

In two types of cost settings, we collect the number of data points in MIMIC-IV. A feature is consider to be static if there is only one point in more than 90\% sequences. For static features, only the first feature update will cause a cost. In \texttt{simple\_cost} setting, each feature update will produce a unit cost except static features. Static feature will produce a unit cost only in the first update. In \texttt{complex\_cost}, we first estimate the cost for each test in real world. Since there are many considerable factors (financial cost, harm to patients, time consumption, etc.), we roughly create five cost standards: basic demographic features have cost 1-2 because each admission will produce a basic cost. Features come from basic measurements have a cost of 2. Observations of ventilator or bedside monitor outputs have a cost of 3-5 because their frequencies are relatively high. Blood gas test results have a cost of 5-10 since they cause a harm to patient and relatively expensive. Other features which use complex measurements or special methods have different costs. The cost for each test are shown in the following table:

\begin{itemize}
    \item \textbf{Basic measurements} BMI (kg/m2):5, Height:2, Weight:2, BMI:5, systolic pressure:5, gender:1, diastolic pressure:5, Height (Inches):2, Weight (Lbs):2, Daily Weight:2, Temperature Fahrenheit:3, Admission Weight (lbs.):2, 
    \item \textbf{Bedside monitor} Heart Rate:3, Respiratory Rate:3, O2 saturation pulseoxymetry:3, Non Invasive Blood Pressure systolic:3, Non Invasive Blood Pressure diastolic:3, Non Invasive Blood Pressure mean:3
    \item \textbf{Ventilation} O2 Flow:5, Inspired O2 Fraction:5, PEEP set:5, Tidal Volume (set):5, Tidal Volume (observed):3, Respiratory Rate (Set):5, Peak Insp. Pressure:5, Plateau Pressure:5, Mean Airway Pressure:5, Total PEEP Level:5, Inspiratory Time:5, Expiratory Ratio:5, Inspiratory Ratio:5, Paw High:5, Vti High:5, Fspn High:5, Apnea Interval:5, Ventilator Tank 1:10, Ventilator Tank 2:10, Minute Volume:5, PSV Level:5, Tidal Volume (spontaneous):3, Cuff Pressure:3
    \item \textbf{Blood gas test} Glucose finger stick (range 70-100):5, Arterial O2 pressure:10, Arterial CO2 Pressure:10, PH (Arterial):10, Arterial Base Excess:10, TCO2 (calc) Arterial:10, Hemoglobin:5, Ionized Calcium:10, Lactic Acid:10, Sodium (whole blood):5, Chloride (whole blood):5, Glucose (whole blood):5, Hematocrit (whole blood - calc):5, Potassium (whole blood):5, Hematocrit (serum):5, WBC:5, Platelet Count:5, Prothrombin time:5, PTT:5, INR:5, Fibrinogen:5, Chloride (serum):5, Creatinine (serum):5, Sodium (serum):5, BUN:5, Anion gap:5, Potassium (serum):5, HCO3 (serum):5, Arterial O2 Saturation:10, Magnesium:5, Glucose (serum):5, Calcium non-ionized:5, Phosphorous:5,Differential-Basos:5, Differential-Eos:5, Differential-Lymphs:5, Differential-Monos:5, Differential-Neuts:5
    \item \textbf{Others} Arterial Blood Pressure systolic:10, Arterial Blood Pressure diastolic:10, eGFR:15, Arterial Blood Pressure mean:10, Central Venous Pressure:10, Respiratory Rate (spontaneous):3, Respiratory Rate (Total):3, PPD:15, MAP:5, shock index:5, sofa score:10
\end{itemize}

\subsection{Evaluation and ablation results} \label{subsec: eval-results}

The main experiment's results are shown in \autoref{tab:exp_detail}. We apply our method with the best predictor in baseline (Catboost in P/F prediction task, LSTM in ventilation task). Ablation results are shown in \autoref{tab:ablation_detail}.

\begin{table}
    \centering
    \begin{tabular}{cccccc}
        cost/loss & GBDT Baseline & LSTM Baseline & Lasso Baseline & SVM-L1 Baseline & Ours \\
        \hline\multirow{6}{*}{pf simple}
        &74.88/60.068&74.88/73.985&74.88/100.576&-/-&59.42/60.073 \\
        &8.20/60.979&8.20/74.513&8.20/100.861&-/-&9.64/60.689 \\
        &4.10/61.465&4.10/76.927&4.10/75.027&-/-&4.89/60.452 \\
        &3.06/63.329&3.06/80.775&2.06/75.750&-/-&2.96/61.526 \\
        &2.04/64.893&2.04/81.956&1.04/76.405&-/-&1.94/61.816 \\
        &1.02/72.082&1.02/81.979&1.02/77.440&-/-&0.89/62.982 \\
        \hline\multirow{6}{*}{pf complex}
        &51.70/60.125&51.70/74.094&51.70/100.575&-/-&40.53/59.983 \\
        &10.15/60.651&10.13/74.431&10.09/98.351&-/-&9.28/60.291 \\
        &2.01/61.313&2.02/75.172&5.07/100.724&-/-&1.84/61.808 \\
        &0.99/63.708&0.99/80.028&3.02/74.935&-/-&0.93/63.893 \\
        &0.70/68.600&0.70/78.248&2.00/74.989&-/-&0.65/65.015 \\
        &0.40/70.538&0.40/80.224&0.98/75.338&-/-&0.37/65.885 \\
        \hline\multirow{6}{*}{vent simple}
        &77.15/0.257&77.15/0.246&-/-&77.15/0.376&56.10/0.243 \\
        &10.24/0.281&10.24/0.262&-/-&9.24/0.403&7.14/0.246 \\
        &5.12/0.303&5.12/0.264&-/-&4.12/0.456&4.35/0.249 \\
        &3.07/0.334&3.07/0.264&-/-&2.07/0.480&2.92/0.244 \\
        &2.05/0.354&2.05/0.264&-/-&2.05/0.480&1.86/0.246 \\
        &1.02/0.401&1.02/0.270&-/-&1.02/0.513&1.09/0.258 \\
        \hline\multirow{6}{*}{vent complex}
        &61.49/0.256&61.49/0.247&-/-&61.49/0.376&41.55/0.246 \\
        &10.31/0.263&10.31/0.253&-/-&10.27/0.390&8.11/0.247 \\
        &5.15/0.271&5.15/0.259&-/-&5.13/0.393&4.50/0.244 \\
        &3.10/0.282&3.09/0.260&-/-&3.08/0.409&2.78/0.248 \\
        &2.05/0.289&2.07/0.261&-/-&2.06/0.414&1.88/0.248 \\
        &1.03/0.316&1.03/0.267&-/-&1.01/0.441&1.29/0.256 \\
        \hline
    \end{tabular}
    \caption{Experiment details. \textbf{pf:} P/F ratio prediction task. Loss=MAE. \textbf{vent:} Ventilation termination prediction. Loss=1-AUROC}
    \label{tab:exp_detail}
\end{table}

\begin{table}
    \centering
    \begin{tabular}{cccccc}
        cost/loss & fixed & w/o norm & w/o predictor update & w/o gate function  & w/o baseline \\
        \hline\multirow{6}{*}{pf}
        &58.95/59.974&60.75/60.014&59.15/60.097&25.60/60.055&56.35/60.135 \\
        &9.49/60.673&9.84/60.796&9.58/60.746&4.55/61.006&9.19/62.589 \\
        &4.54/61.243&5.06/61.076&5.07/62.114&1.99/62.302&4.71/64.143 \\
        &2.74/62.127&2.87/61.623&2.92/62.420&1.39/62.865&2.63/65.052 \\
        &1.83/63.289&1.92/62.257&1.86/65.062&0.26/67.311&1.97/64.685 \\
        &0.95/64.797&1.04/62.786&0.98/64.670&0.12/73.217&0.88/67.617 \\
        \hline\multirow{6}{*}{vent}
        &55.18/0.247&54.95/0.251&55.69/0.290&17.32/0.249&-/- \\
        &9.16/0.248&5.05/0.245&6.08/0.289&2.75/0.246&-/- \\
        &3.98/0.246&4.94/0.246&4.22/0.284&2.50/0.246&-/- \\
        &2.62/0.246&2.63/0.245&2.66/0.279&1.91/0.246&-/- \\
        &1.94/0.249&1.92/0.245&1.89/0.278&1.88/0.247&-/- \\
        &1.14/0.257&1.07/0.258&1.03/0.303&0.83/0.265&-/- \\
        \hline
    \end{tabular}
    \caption{Ablation results. Fixed: the cost coefficient is fixed. w/o norm: No prediction reward normalization. w/o prediction update: predictor will not be updated when policy is converged. w/o gate multiplier: No gate control for cost coefficient. w/o baseline: Prediction reward will not be normalized by baseline predictor's performance.  '-/-' means we did not use baseline in ventilation prediction task.}
    \label{tab:ablation_detail}
\end{table}
\end{document}